\documentclass[runningheads]{llncs}
\usepackage[T1]{fontenc}
\usepackage{graphicx}
\usepackage[utf8]{inputenc} 
\usepackage[T1]{fontenc}    
\usepackage{pifont}
\usepackage[colorlinks,
    linkcolor=red,citecolor=citecolor,urlcolor=blue]{hyperref}       
\usepackage{graphicx}
\usepackage{xcolor}
\usepackage{todonotes}
\usepackage[table,xcdraw]{xcolor}
\usepackage{amssymb}
\usepackage{threeparttable}
\usepackage{multirow}
\usepackage{amsmath}
\usepackage{bm}
\usepackage{float}
\usepackage{soul}
\usepackage[noend]{algpseudocode}
\usepackage{enumitem} 
\usepackage[subrefformat=parens,labelformat=parens]{subfig}

\usepackage{wrapfig}
\usepackage{xspace}

\usepackage{multirow}
\usepackage[normalem]{ulem}
\useunder{\uline}{\ul}{}

\definecolor{citecolor}{RGB}{34,139,34}
\definecolor{mydarkblue}{rgb}{0,0.08,1}
\definecolor{mydarkgreen}{rgb}{0.02,0.6,0.02}
\definecolor{mydarkred}{rgb}{0.8,0.02,0.02}
\definecolor{mydarkorange}{rgb}{0.40,0.2,0.02}
\definecolor{mypurple}{RGB}{111,0,255}
\definecolor{myred}{rgb}{1.0,0.0,0.0}
\definecolor{mygold}{rgb}{0.75,0.6,0.12}
\definecolor{myblue}{rgb}{0,0.2,0.8}
\definecolor{mydarkgray}{rgb}{0.,0.2,0.2}

\definecolor{lightred}{RGB}{255,235,235}
\definecolor{lightgreen}{RGB}{235,255,235}
\definecolor{lightblue}{RGB}{235,235,255}
\definecolor{lightcyan}{RGB}{235,255,255}
\definecolor{lightmagenta}{RGB}{255,235,255}
\definecolor{lightyellow}{RGB}{255,255,235}

\definecolor{qxkcolor}{RGB}{215,235,255}
\definecolor{softmaxcolor}{RGB}{230,235,255}
\definecolor{probxvcolor}{RGB}{255,255,235}

\definecolor{topkcolor}{RGB}{255,235,235}
\definecolor{zecolor}{RGB}{255,255,235}
\definecolor{dynacolor}{RGB}{235,255,255}

\definecolor{reviewcolor}{RGB}{0,0,200}

\newcommand{\calD}{\mathcal{D}}

\newcommand{\calL}{\mathcal{L}}

\algdef{SE}[DOWHILE]{Do}{doWhile}{\algorithmicdo}[1]{\algorithmicwhile\ #1}%

\newcommand{\squishlist}{
 \begin{list}{$\bullet$}
  { \setlength{\itemsep}{0pt}
     \setlength{\parsep}{3pt}
     \setlength{\topsep}{3pt}
     \setlength{\partopsep}{0pt}
     \setlength{\leftmargin}{1.5em}
     \setlength{\labelwidth}{1em}
     \setlength{\labelsep}{0.5em} } }
     
\newcommand{\squishend}{
  \end{list}  }     

\newcommand{\name}{\texttt{CAMO}\xspace}

\begin{document}
\title{\name: \underline{C}ausality-Guided \underline{A}dversarial \underline{M}ultimodal D\underline{O}main Generalization for Crisis Classification}
%
%
\author{Pingchuan Ma\inst{1} \and
Chengshuai Zhao\inst{1} \and
Bohan Jiang\inst{1} \and
Saketh Vishnubhatla\inst{1} \and
Ujun Jeong\inst{1} \and
Alimohammad Beigi\inst{1} \and
Adrienne Raglin\inst{2} \and
Huan Liu\inst{1}
}
\authorrunning{P. Ma et al.}
%
\institute{Arizona State University, Tempe AZ 85281, USA \and
DEVCOM Army Research Laboratory, Adelphi MD 20783, USA
}
%
\maketitle              
\begin{abstract}
\label{abstract}
Crisis classification in social media aims to extract actionable disaster-related information from multimodal posts, which is a crucial task for enhancing situational awareness and facilitating timely emergency responses. However, the wide variation in crisis types makes achieving generalizable performance across unseen disasters a persistent challenge. Existing approaches primarily leverage deep learning to fuse textual and visual cues for crisis classification, achieving numerically plausible results under in-domain settings. However, they exhibit poor generalization across unseen crisis types because they \ding{202}~do not disentangle spurious and causal features, resulting in performance degradation under domain shift, and \ding{203}~fail to align heterogeneous modality representations within a shared space, which hinders the direct adaptation of established single-modality domain generalization (DG) techniques to the multimodal setting.
To address these issues, we introduce a causality-guided multimodal domain generalization (MMDG) framework that combines adversarial disentanglement with unified representation learning for crisis classification. The adversarial objective encourages the model to disentangle and focus on domain-invariant causal features, leading to more generalizable classifications grounded in stable causal mechanisms. The unified representation aligns features from different modalities within a shared latent space, enabling single-modality DG strategies to be seamlessly extended to multimodal learning. Experiments on the different datasets demonstrate that our approach achieves the best performance in unseen disaster scenarios. 
\end{abstract}

\section{Introduction}
Crisis classification in social media aims to classify multimodal posts to extract actionable disaster information that supports timely emergency response~\cite{Crisis_abavisani_CVPR2020}. 
With the rapid growth of social platforms, affected individuals and eyewitnesses frequently share first-hand reports, images, and situational updates during unfolding events, making social media a vital channel for real-time crisis management~\cite{Crisis_xie_IPM26,Crisis_etal_EMNLP23,Crisis_Rezk_Access23}. 
Effectively analyzing this multimodal data enables humanitarian organizations to enhance situational awareness, prioritize rescue efforts, and allocate resources more efficiently.
\begin{figure*}
\vspace{-10pt}
    \centering
    \vspace{-5pt}
    \subfloat[]{\includegraphics[width=0.45\textwidth]{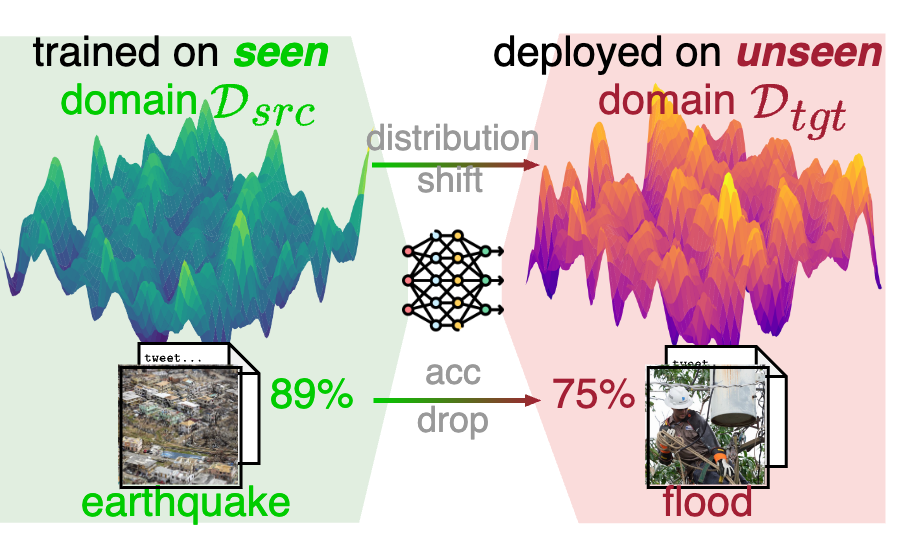}
    \label{fig:DG_drop}
    }
    \hspace{5pt}
    \subfloat[]{\includegraphics[width=0.48\textwidth]{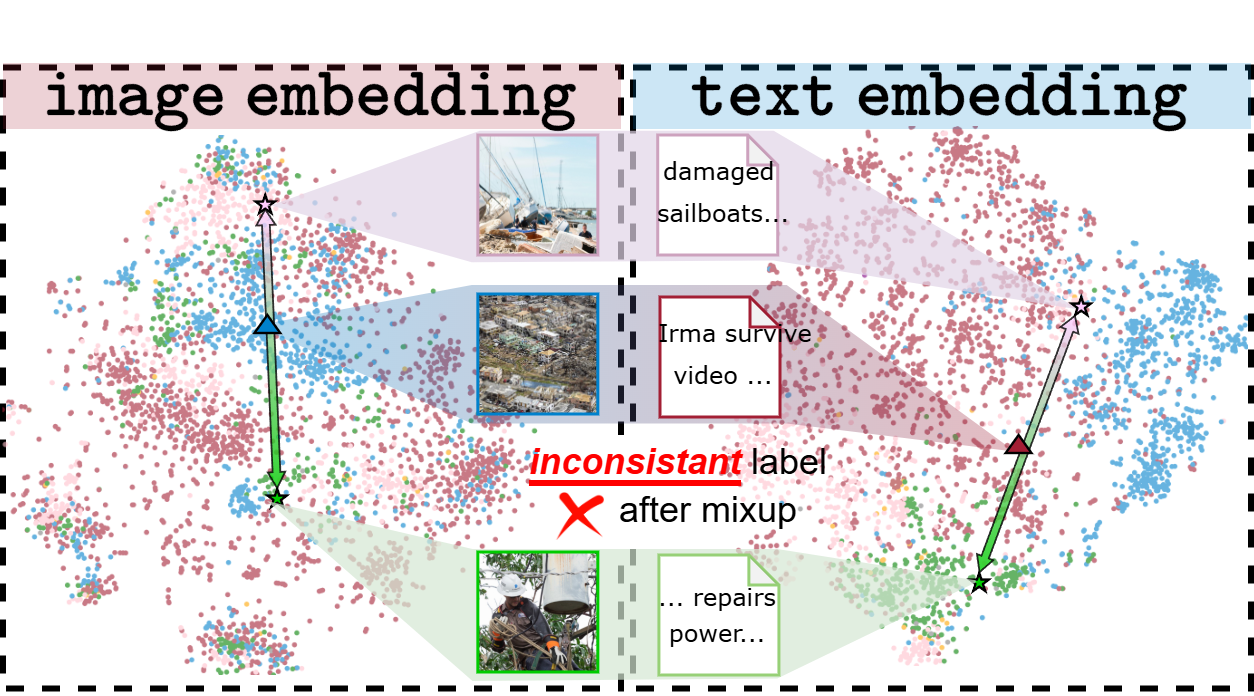}
    \label{fig:MixupFailure}
    }
    \vspace{-5pt}
    \caption{(a) Model performance drops significantly when deployed on unseen disaster types.
    (b) Direct apply single modality DG techniques is not feasible in MMDG problem. For example, Mix-up produces inconsistent labels.
    }
    \label{fig:motivation}
    \vspace{-20pt}
\end{figure*}


Despite advances in multimodal deep learning for crisis information classification, most existing models assume that the crisis types encountered during deployment are similar to those seen during training, achieving numerically plausible performance under this assumption~\cite{Crisis_Mandal_WWW24,Crisis_abavisani_CVPR2020}.
In practice, each crisis presents distinct visual patterns, linguistic expressions, and contextual cues.
This mismatch leads to severe performance degradation when models are deployed in real-world scenarios where unseen crises may occur, as shown in Figure.~\ref{fig:DG_drop}.
Collecting and annotating sufficient multimodal data for every new event is infeasible, especially under time-critical conditions.
Therefore, developing models with strong domain generalization (DG) to unseen crisis events is crucial for achieving reliable and adaptive crisis information analysis in real-world response systems.

While recent progress in domain generalization has improved the robustness of crisis classification models~\cite{Crisis_xie_IPM26}, two major challenges remain.
\ding{202}~Current methods often learn representations that mix domain-invariant causal features which determine true crisis-related categories with spurious correlations tied to specific domains, such as background textures, linguistic styles, or data collection artifacts.
When the crisis context shifts, these spurious cues fail to transfer, causing significant degradation on unseen events.
\ding{203}~Multimodal crisis classification models typically encode visual and textual inputs in heterogeneous representation spaces without explicit alignment, causing the two modalities to generalize inconsistently across domains.
This misalignment prevents conventional single-modality domain generalization techniques from being directly extended to multimodal settings, as illustrated in Fig.~\ref{fig:MixupFailure}, where mix-up in representation space serves as an example of such failure~\cite{MMDG_hai_ICCV25}.

To address these challenges, we formulate crisis classification as a multimodal domain generalization problem and propose \name, a framework consists of \ding{202}~\uline{\textit{a causality-guided adversarial disentanglement module}} that separates domain-invariant causal features from spurious domain-specific cues, encouraging the model to rely on stable mechanisms that transfer across unseen crisis types.
\ding{203}~\uline{\textit{a unified representation learning module}} that disentangles and aligns the modal-general factors from visual and textual features into a shared latent space, enabling established single-modality domain generalization strategies to be seamlessly adapted to the multimodal setting.
Together, these components enhance robustness to domain shift and improve reliability on unseen crisis types, contributing to more effective crisis monitoring and response.
Our main contributions are summarized as follows:
\squishlist
{\item We introduce the causal graph underlying the crisis classification process and propose a causality-guided disentanglement framework for domain generalization through adversarial learning.}
{\item We develop a unified representation across modalities, allowing single-modality domain generalization strategies to be extended to multimodal learning.}
{\item We achieved 4\% - 21\% performance improvement on CrisisMMD~\cite{Crisis_alam_AAAI18} and DMD~\cite{Crisis_mouzannar_ISCRAM18}, outperforming existing crisis classification and multimodal domain generalization methods.}
\squishend
\section{Related Work}
\label{sec:Background}
\subsection{Machine Learning for Disaster Classification}
Recent studies have applied machine learning to crisis classification using either single~\cite{Crisis_jain24_ICIC3S24} or multimodal~\cite{Crisis_abavisani_CVPR2020,Crisis_Mandal_WWW24,Crisis_etal_EMNLP23,Crisis_Rezk_Access23} data, achieving numerically plausible results under the in-distribution assumption.
Several studies have further explored temporal domain generalization, treating past and future disasters as separate domains~\cite{Crisis_abavisani_CVPR2020,Crisis_Mandal_WWW24}.
However, these studies overlook the ability to generalize across different disaster types, which is arguably more critical due to the larger domain gaps and the limited variety of disaster categories available in existing datasets.
\subsection{Causal Representation Learning for Domain Generalization}
Invariant causal mechanism has been utilized to improve generalization in different domains, such as hate speech detection, image classification and sentiment analysis.
For example, CATCH~\cite{DG_paras_WSDM24} proposed a cross-platform hate speech detection framework that disentangles features into platform-specific and platform-general components using a variational autoencoder.
CIRL~\cite{DG_fangrui_CVPR22} augmented the frequency domain of input images to encourage models to learn causal features invariant to superficial correlations.
Wang et al.~\cite{DG_siyin_LREC_COLING24} introduced a backdoor adjustment-based causal model that separates domain-specific and domain-invariant representations through a dual-encoder structure with causal regularization losses.
However, these causal representation learning methods are designed for single-modality tasks and cannot effectively address multimodal scenarios.
\subsection{Multi-modal Domain Generalization}
Recently, an increasing number of studies have focused on multimodal domain generalization (MMDG).
CMRF~\cite{MMDG_fan_NIPS24}, which integrates Sharpness-Aware Minimization with Simple Moving Average, addresses modality competition by interpolating toward a shared flat minimum in the loss landscape.
OGM-GE~\cite{MMDG_peng_CVPR22} mitigates optimization imbalance by monitoring and adaptively scaling gradients across modalities.
SimMMDG~\cite{MMDG_dong_NIPS23} improves generalization through supervised contrastive learning with cross-modal translation regularization, which also alleviates the modality-missing problem.
Huang et al.~\cite{MMDG_hai_ICCV25} further explored unified representation learning to enable the direct use of single-modality DG techniques in multimodal settings.
Despite these advances, existing studies mainly emphasize optimization and representation alignment while neglecting the causal factors behind domain invariance. In contrast, our framework explicitly disentangles causal and spurious features to promote generalization grounded in stable causal mechanisms.
\section{Method \name}
We denote the set of source domains as $\mathcal{D}_{\text{src}} = \{D_1, \dots, D_n\}$, 
where each domain $D_i$ contains multimodal samples $(x^v, x^t, y)$ with image $x^v$, 
text $x^t$, and corresponding label $y$. 
At test time, the model is evaluated on an unseen target domain $D_{\text{tgt}}$, 
whose visual and textual distributions differ from those in $\mathcal{D}_{\text{src}}$. 
Our objective is to learn a classifier $f: (x^v, x^t) \rightarrow y$ that 
predicts $y$ reliably across unseen domains.

To meet this objective, \name ~ \ding{202} disentangles domain-invariant causal features 
from domain-specific spurious factors via adversarial disentanglement, 
and \ding{203} learns a unified representation that aligns modal-general 
features. 
This design both restricts the classifier to stable causal cues 
and enables the seamless adaption of single-modality DG techniques in the multimodal setting. 
\subsection{\name Overview}
\begin{figure}
\centering
\vspace{-15pt}
\includegraphics[width=0.8\textwidth]{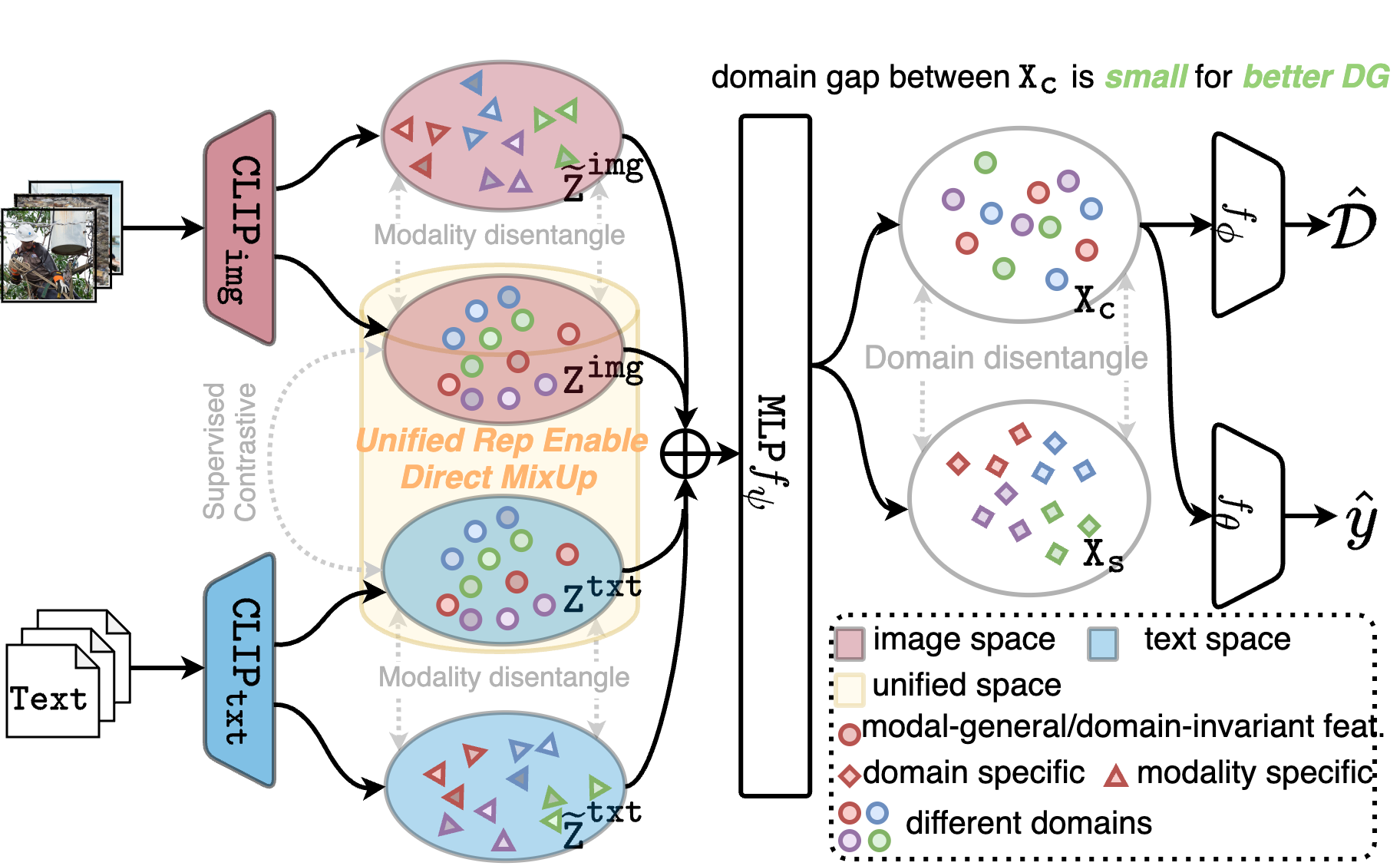}
\caption{\name framework consists of (1) unified representation learning that disentangles and aligns modal-general features for direct use of DG methods, and (2) causality-guided adversarial disentanglement that isolates domain-invariant features to enhance domain generalization.} 
\vspace{-15pt}
\label{fig:main_framework}
\end{figure}
Figure~\ref{fig:main_framework} illustrates the \name framework. 
For each post, the visual and textual modalities are first encoded by their corresponding pretrained feature extractors, producing modality-specific embeddings $f^m$. 
These embeddings are then projected into two subspaces, a modality-specific representation $\widetilde{z}^m$ and a modality-general representation $z^m$, through modality-dependent projection heads:
\[
[\widetilde{z}^m; z^m] \leftarrow \mathrm{Proj}^m(f^m),
\]
where $m$ indexes the modality. 
The representations from all modalities are concatenated and passed through a multilayer perceptron $f_{\psi}(\cdot)$ to disentangle the latent representation into a domain-specific component $X_s$ and a domain-invariant causal component $X_c$. 
A discriminator $f_{\phi}(\cdot)$ predicts the domain label $\hat{D_l} = f_{\phi}(X_c)$, 
while a classification head $f_{\theta}(\cdot)$ produces the final category prediction $\hat{y} = f_{\theta}(X_c)$. 
This design enforces the causal feature $X_c$ to capture invariant semantics across domains while free from spurious correlations in $X_s$.
\subsection{Causally Guided Disentanglement for Domain Generalization via Adversarial Learning}
\begin{wrapfigure}[14]{rH}{0.3\textwidth}
\begin{minipage}{0.3\textwidth}
    \centering
    \vspace{-17pt}
    \includegraphics[width=0.99\columnwidth]{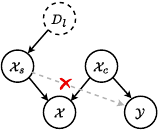}
    \caption{Causal graph showing the data-generating mechanism in disaster classification problem.}
    \vspace{-20pt}
    \label{fig:CausalGraph}
\end{minipage}
\end{wrapfigure}
Figure.~\ref {fig:CausalGraph} depicts the structural relationships underlying disaster-related social media posts. 
The domain variable $D_l$ serves as the parent of the domain-specific factors $X_s$, indicating that different domains induce distinct distributions over these variables, including variations in visual background, linguistic style, and contextual framing. 
Importantly, $X_s$ is not causally responsible for the label $y$; rather, it can become associationally correlated with $y$ through dataset bias or statistical dependence mediated by the observed variable $X$ or its correlation with $X_c$. 
In contrast, the domain-invariant features $X_c$ exhibit a causal correlation with $y$, as $X_c$ encodes stable semantic cues that consistently determine the label across domains. 
The observed multimodal post $X$ is generated jointly from both $X_s$ and $X_c$, following the structural causal model (SCM):
\[
X_s \leftarrow f_s(D_l, \epsilon_s), \quad 
X_c \leftarrow f_c(\epsilon_c), \quad 
X \leftarrow f_x(X_s, X_c, \epsilon_x), \quad 
y \leftarrow f_y(X_c, \epsilon_y),
\]
where $\epsilon$ represents exogenous noise variables. 
Under domain shifts, the conditional distribution $P(X_s \mid D_l)$ varies across domains, whereas the causal mechanism $P(y \mid X_c)$ remains invariant. 
Consequently, robust domain generalization requires disentangling the causal signal $X_c$ from the domain-dependent spurious component $X_s$, ensuring that predictions depend exclusively on the invariant causal pathway $X_c \rightarrow y$ rather than on incidental correlations.

An essential observation derived from the causal graph is the \textbf{independence relation
$X_c \perp D_l$}, which states that the invariant causal features should be independent of the domain variable. 
This property ensures that the predictive mechanism $P(y \mid X_c)$ remains stable across domains, 
providing the foundation for generalization under distribution shift. 
Such an independence condition can be implicitly encouraged through an adversarial objective, 
where a discriminator attempts to identify the domain label $D_l$ from the learned invariant representation,  while the feature extractor aims to remove domain information that violates this independence. 
This formulation naturally motivates the use of adversarial learning to disentangle $X_c$ from $X_s$, 
yielding a representation that preserves the causal semantics relevant for prediction 
while suppressing domain-dependent spurious signals.

The causality guided adversarial learning is performed by alternating between training discriminator $f_{\phi}$ and the invariant causal feature learning. Specifically, $f_{\phi}$ is first trained through minimizing cross entropy loss while all other parameters in \name frozen:
\begin{equation}
    \calL_{disc} = CE(f_{\phi}(x_c), D_l)
\end{equation}
In the second step, the invariant causal feature is learned through learning to correctly classify the task label and confuse the discriminator during which the discriminator is frozen.
\begin{equation}
    \calL_{task} = CE(f_{\theta}(x_c), y) + \gamma \cdot CE(f_{\phi}(x_c), D_l)
\end{equation}
where $\gamma$ is the multiplier in the gradient reverse layer.
\subsection{Unified Representation Enables Domain Generalization in Multimodal Learning}
\label{sec:UR_learning}
Different modalities inherently reside in heterogeneous representation spaces, 
making it infeasible to directly apply single-modality domain generalization methods to the multimodal setting. 
For instance, textual and visual encoders capture distinct semantic structures and statistical characteristics, 
which may generalize along inconsistent directions when trained independently. 
To enable unified generalization behavior across modalities, it is essential to align their representations into a shared latent space, 
where semantically related instances exhibit proximity regardless of modality. 
However, not all semantic information is shared, each modality contains both general and modality-specific factors. 
Therefore, the unified representation must preserve modality-general semantics while disentangling modality-specific components that do not transfer across modalities. 

We achieve this disentanglement through supervised contrastive learning\cite{ML_khosla_NIPS_20}, 
which explicitly pulls together samples with shared semantic labels and pushes apart those from different categories. In a batch of $N$ posts and each post has $M$ modalities, $i \in I \equiv\{1, \ldots, M \times N\}$ is the index for one modality of one sample. We define $A(i) \equiv I \backslash\{i\}, P(i) \equiv\left\{p \in A(i): y_p=y_i\right\}$ as the indices consists of all positive samples in the batch sharing the sample label as $i$. $|P(i)|$ stands for its cardinality. Then, the unified representation is learned through minimizing the following supervised contrastive loss, where $\tau \in \mathbb{R}^+$ is the temperature:
\begin{equation}
    \mathcal{L}_{supcon}=\sum_{i \in I} \frac{-1}{|P(i)|} \sum_{p \in P(i)} \log \frac{\exp \left(\boldsymbol{z}_i \cdot \boldsymbol{z}_p / \tau\right)}{\sum_{a \in A(i)} \exp \left(\boldsymbol{z}_i \cdot \boldsymbol{z}_a / \tau\right)}
\end{equation}
Based on this unified representation, single-modality domain generalization techniques can be effectively applied in the multimodal setting, 
enhancing robustness and consistency across unseen crisis domains. Specifically, in \name we apply Mixup by replacing the modal-general feature $z^{m}$ and label $y$ with their mixed counterparts $\bar{z}^{m}$ and $\bar{y}$ defined below, where $\lambda$ is sampled from a beta distribution.
Mixing is performed within the same domain, so the corresponding domain label $\mathcal{D}_l$ remains unchanged:
\begin{equation}
    \bar{\mathbf{z}}_i^m=\lambda \mathbf{z}_i^m+(1-\lambda) \mathbf{z}_j^m, \bar{\mathbf{y}}=\lambda \mathbf{y}_i+(1-\lambda) \mathbf{y}_j
\end{equation}
\subsection{Model Training}
To further regularize the training, we use orthogonal loss between domain-invariant and specific features and modal-general and specific features using cosine similarity:
\begin{equation}
    \calL_{\perp} = \frac{x_c \cdot x_s}{|x_c||x_s|} + \sum_{m\in\mathcal{M}}\frac{z^m\cdot \widetilde{z}^m}{|z^m||\widetilde{z}^m|}
\end{equation}
The total losses in the alternating steps are:
\begin{equation}
    \calL_{step1} = \calL_{disc},\quad \calL_{step2} = \alpha_1\calL_{task} + \alpha_2\calL_{supcon} + \alpha_3\calL_{\perp}
\end{equation}
\section{Experiment Results}
\label{sec:Experiment}

\subsection{Experimental Setup}
\subsubsection{Benchmarks}
We evaluate our method on the only two multimodal crisis benchmark available, CrisisMMD\cite{Crisis_alam_AAAI18} and DMD\cite{Crisis_mouzannar_ISCRAM18}, which both contain paired image-text samples collected from social media posts during real-world crisis events.
\paragraph{CrisisMMD} aggregates data from diverse disaster events categorized into four broad domains: earthquakes $\calD_{eq}$, hurricanes $\calD_{hu}$, floods $\calD_{fl}$, and wildfires $\calD_{wf}$. The dataset originally defines three annotation tasks: (1) binary informativeness classification, which distinguishes relevant disaster-related posts from non-informative ones; (2) 8-way humanitarian categorization, which classifies posts into fine-grained classes such as infrastructure damage, vehicle damage, or human injury; and (3) 3-way damage severity assessment, categorizing damage as mild, medium, or severe. In our study, we use only the first two tasks, as the third is a unimodal classification problem.
\paragraph{DMD} was originally a six-way classification benchmark, lacking explicit domain labels required for DG. To construct a DG setting, we reformulate the dataset into a binary classification task with damage and non-damage labels, and designate the five damage-related categories, namely, infrastructural $\calD_{infra}$, natural landscape $\calD_{nature}$, fires $\calD_{fires}$, floods $\calD_{flood}$, and human damage $\calD_{human}$, as distinct domains. To maintain domain-specific characteristics, the negative samples from the original non-damage class, comprising advertisement, building, car, food, and nature, are exclusively assigned to separate domains. This configuration yields a five-domain benchmark suitable for evaluating DG performance.
\subsubsection{Training settings and evaluation metrics} We set $\alpha_1 = 1, \alpha_2=3, \alpha_3=1$, $\tau$ in $\calL_{supcon}$ is set to be 0.7. 
To ensure stable adversarial learning, gradient reverse multiplier $\gamma$ is gradually increased following a cosine schedule during the first 400 iterations to 10.
The model is optimized using Adam with learning rate $0.002$ which decayed to $2e^{-5}$ with a cosine schedule and a batch size of 16. Domain generalization performance is assessed using a leave-one-domain-out evaluation, in which one disaster/damage type is designated as the target domain and the remaining types constitute the source domains; we quantify performance by the classification accuracy on the held-out target.
\subsubsection{Baselines}
To demonstrate the effectiveness and generalization capability of our approach, 
we compare six representative models across three categories: 
(1) \textit{a basic multimodal baseline} that jointly trains on concatenated visual and textual features extracted from VGG\cite{CV_simonyan_ICLR15} and BERT\cite{NLP_devlin_NAACL19}; 
(2) \textit{two models developed for disaster classification}, including SSE-Cross-BERT-DenseNet (SCBD)\cite{Crisis_abavisani_CVPR2020}, which employs cross-modal attention to suppress uninformative or misleading signals from weak modalities, and CLMC\cite{Crisis_Mandal_WWW24}, which replaces VGG and BERT with CLIP\cite{MM_radford_ICML21} to enhance visual–textual alignment; 
and (3) \textit{two recent MMDG methods}, SimMMDG\cite{MMDG_dong_NIPS23} and CMRF\cite{MMDG_fan_NIPS24}. 
SimMMDG aligns representation spaces across modalities, whereas CMRF seeks a shared flat minimum in the loss landscape. 
Neither method explicitly disentangles domain-invariant and domain-specific factors, relying instead on entangled feature representations. For a fair comparison, we reimplemented SimMMDG and CMRF with CLIP, as their original feature extractors for video, audio and optical flow inputs are not suitable for our setting.
\subsection{Main Results}
\begin{table}
\vspace{-10pt}
\centering
\caption{Classification accuracy on the target domain for two multimodal classification tasks from the CrisisMMD dataset, evaluated across six models.}\label{tab:main_result_crisis}
\resizebox{\columnwidth}{!}{
\begin{tabular}{cccccc}
\hline
                     & \multicolumn{5}{c}{\textbf{CrisisMMD-informative}}                                                                                                                                                                                                                                                            \\ \cline{2-6} 
models               & $\calD_{fl}, \calD_{wf}, \calD_{hu}   \rightarrow \calD_{eq}$ & $\calD_{fl}, \calD_{wf}, \calD_{eq}   \rightarrow \calD_{hu}$ & $\calD_{fl}, \calD_{eq}, \calD_{hu}   \rightarrow \calD_{wf}$ & $\calD_{eq}, \calD_{wf}, \calD_{hu}   \rightarrow \calD_{fl}$ & avg                  \\ \hline
VGG+BERT             & 0.846                                                             & 0.821                                                             & 0.793                                                             & 0.915                                                             & 0.844                \\
SCBD                 & 0.838                                                             & 0.766                                                             & 0.810                                                             & 0.849                                                             & 0.816                \\
CLMC                 & {\ul \textit{0.882}}                                              & 0.842                                                             & 0.781                                                             & 0.940                                                             & 0.861                \\
SimMMDG               & 0.876                                                             & 0.807                                                             & 0.826                                                             & \textbf{0.969}                                                    & 0.869                \\
CMRF                 & \textbf{0.886}                                                    & {\ul \textit{0.859}}                                              & {\ul \textit{0.878}}                                              & 0.943                                                             & {\ul \textit{0.891}} \\
\name & 0.867                                                             & \textbf{0.898}                                                    & \textbf{0.897}                                                    & {\ul \textit{0.949}}                                              & \textbf{0.903}       \\ \cline{2-6} 
                     & \multicolumn{5}{c}{\textbf{CrisisMMD-humanitarian}}                                                                                                                                                                                                                                                           \\ \cline{2-6} 
VGG+BERT             & 0.726                                                             & 0.655                                                             & 0.697                                                             & \textbf{0.935}                                                    & 0.753                \\
SCBD                 & 0.725                                                             & 0.693                                                             & 0.738                                                             & 0.906                                                             & 0.766                \\
CLMC                 & \textbf{0.751}                                                    & \textbf{0.861}                                                    & 0.752                                                             & 0.929                                                             & \textbf{0.823}       \\
SimMMDG               & 0.727                                                             & 0.654                                                             & 0.756                                                             & 0.936                                                             & 0.768                \\
CMRF                 & 0.700                                                             & 0.658                                                             & {\ul \textit{0.789}}                                              & 0.931                                                             & 0.769                \\
\name & {\ul \textit{0.729}}                                              & {\ul \textit{0.736}}                                              & \textbf{0.804}                                                    & {\ul \textit{0.934}}                                              & {\ul \textit{0.801}} \\ \hline
\end{tabular}
}
\vspace{-10pt}
\end{table}

\begin{table}
\vspace{-10pt}
\centering
\caption{Classification accuracy on the target domain for two multimodal classification tasks from the DMD dataset, evaluated across six models.}\label{tab:main_result_dmd}
\begin{tabular}{ccccccc}
\hline
                     & \multicolumn{6}{c}{\textbf{DMD-damge or not}}                                                                                               \\ \cline{2-7} 
models               & $\calD_{infra}$      & $\calD_{nature}$     & $\calD_{fires}$ & $\calD_{flood}$      & $\calD_{human}$      & avg                  \\ \hline
VGG+BERT             & 0.780                & 0.690                & 0.825           & 0.702                & 0.510                & 0.701                \\
SCBD                 & 0.785                & 0.609                & 0.767           & 0.681                & 0.451                & 0.659                \\
CLMC                 & 0.928                & 0.833                & 0.957           & \textbf{0.904}       & {\ul \textit{0.934}} & {\ul \textit{0.911}} \\
SimMMDG               & {\ul \textit{0.937}} & 0.824                & \textbf{0.974}  & {\ul \textit{0.896}} & 0.865                & 0.899                \\
CMRF                 & 0.928                & \textbf{0.842}       & 0.950           & 0.859                & 0.924                & 0.900                \\
\name & \textbf{0.939}       & {\ul \textit{0.837}} & \textbf{0.974}  & 0.892                & \textbf{0.951}       & \textbf{0.919}       \\ \hline
\end{tabular}
\vspace{-20pt}
\end{table}

Table.~\ref {tab:main_result_crisis} shows the classification accuracy on $\calD_{tgt}$ when we set different combinations as $\calD_{src}$ in CrisisMMD, the \textbf{best} and the \uline{\textit{second}} performance is highlighted. \name achieves the best overall performance.
CLMC outperforms VGG+BERT and SCBD, which we attribute to its stronger pretrained feature extractor.
\name further improves on CLMC by explicitly targeting domain-invariant causal features and by addressing modality competition.
By contrast, SimMMDG and CMRF remain vulnerable under distribution shift because they rely on entangled features rather than isolating invariant causal components, causing their performance to decline on unseen disaster domains.


\begin{wrapfigure}[19]{rH}{0.55\textwidth}
\vspace{27pt}
\begin{minipage}{0.55\textwidth}
    \centering
    \vspace{-25pt}
    \includegraphics[width=1\columnwidth]{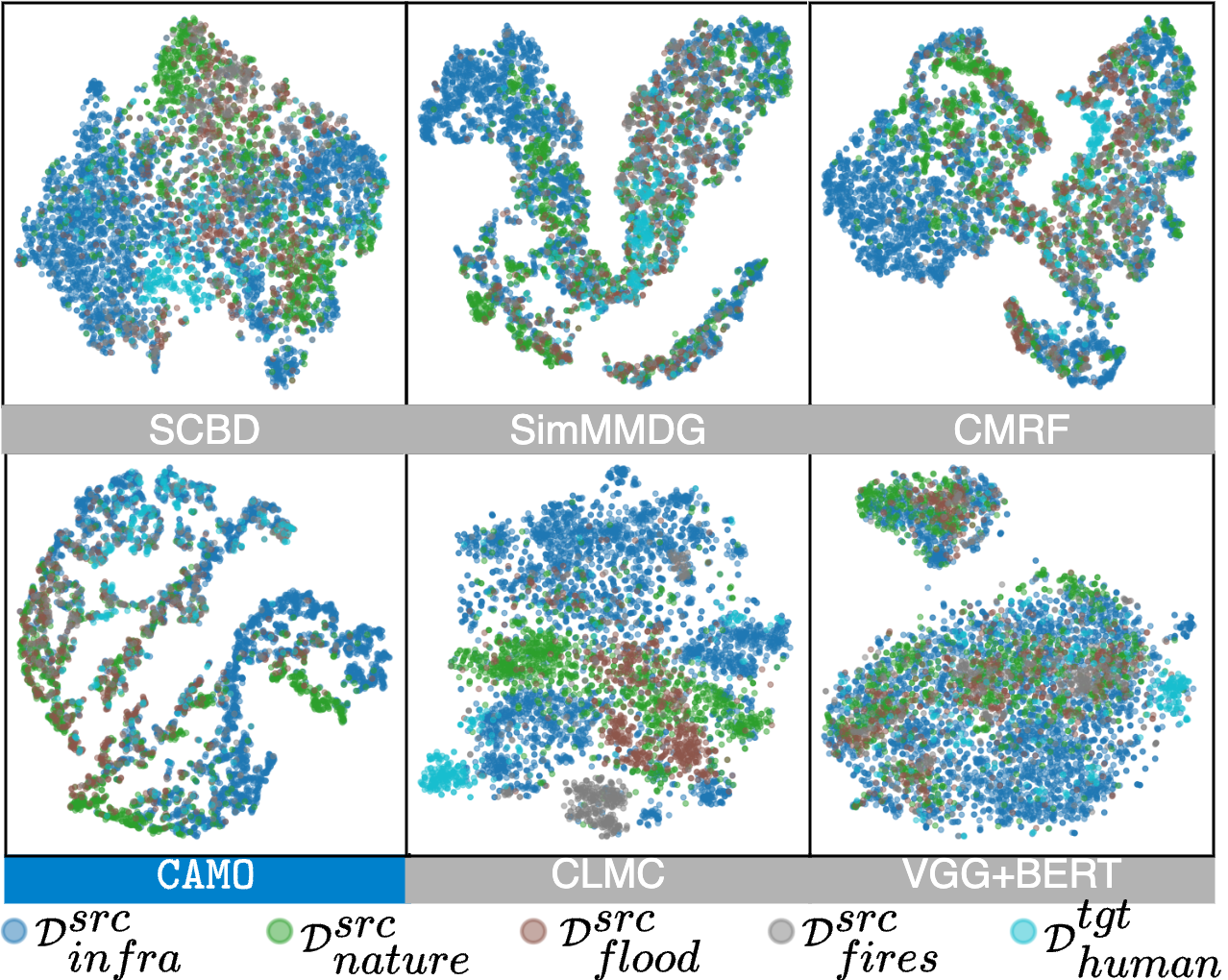}
    \caption{t-SNE visualization of the feature representations used by the baselines and \name. Features obtained by \name exhibit a smaller domain gap across different domains.}
    \label{fig:tSNE_feature}
    \vspace{-20pt}
\end{minipage}
\end{wrapfigure}
Table~\ref{tab:main_result_dmd} presents the classification accuracy on the unseen domain when models are trained on the remaining domains. 
\name consistently achieves the best overall performance among all baselines. 
Although this task resembles the informative classification task in CrisisMMD, the performance of VGG+BERT and SCBD drops noticeably. 
We attribute this decline to the larger domain gap in DMD, where the source and target domains differ more significantly than in CrisisMMD. 
In particular, the negative samples in DMD are intentionally constructed to vary across domains, amplifying the cross-domain discrepancy and making generalization more challenging. Over the two benchmarks, \name outperformed the baselines by 4\% on average and 21\% maximum improvement.
\subsection{Visual Analysis}
\begin{figure}
\centering
\vspace{-25pt}
\includegraphics[width=0.95\textwidth]{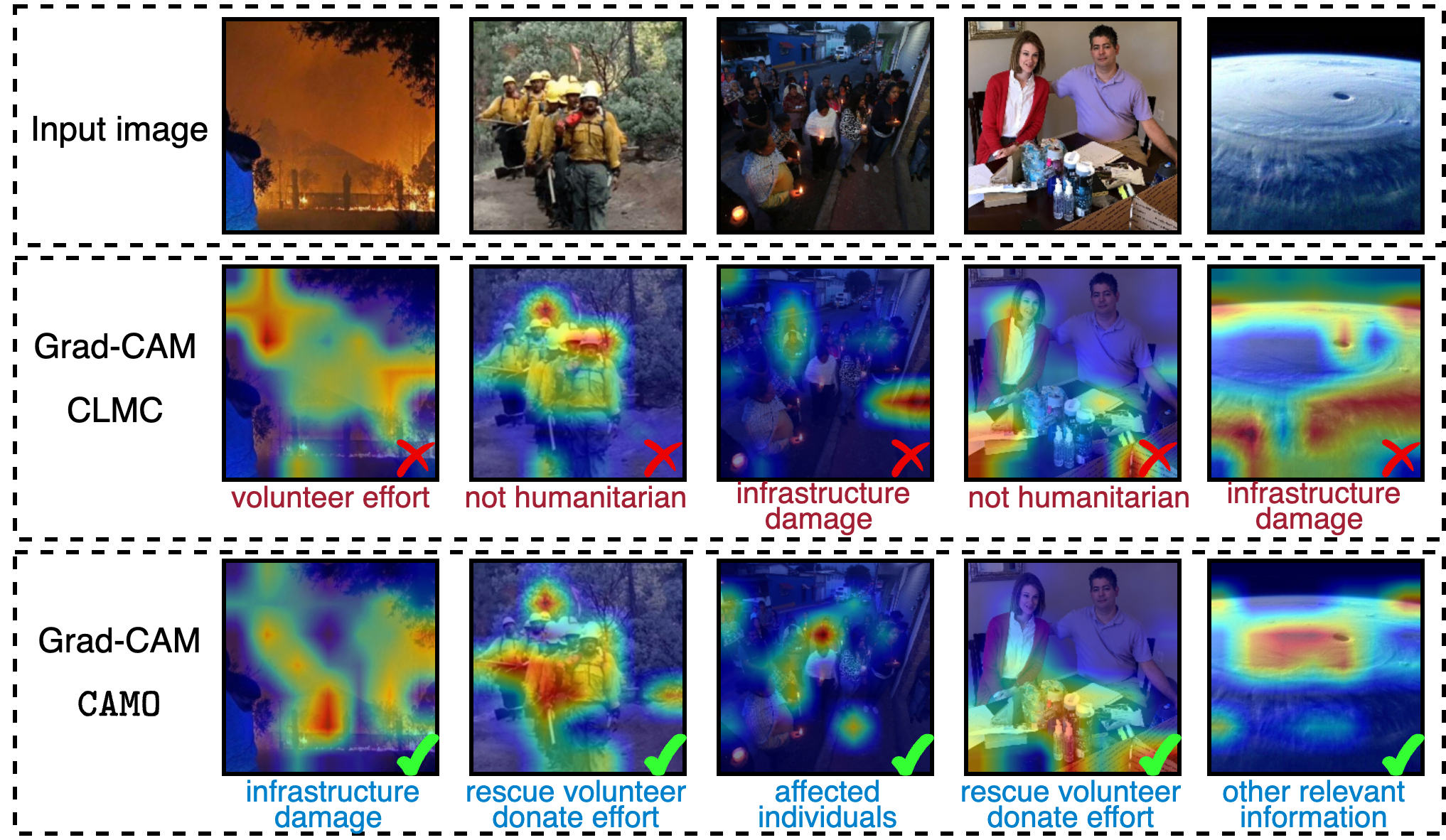}
\caption{Grad-CAM visualization comparing CLMC and \name. While CLMC often focuses on spurious correlations such as background regions, \name consistently attends to causal, semantically relevant areas, showing improved generalization.} 
\vspace{-25pt}
\label{fig:cam}
\end{figure}
\begin{wrapfigure}[18]{rH}{0.4\textwidth}
\begin{minipage}{0.4\textwidth}
    \centering
    \vspace{-20pt}
    \includegraphics[width=1\columnwidth]{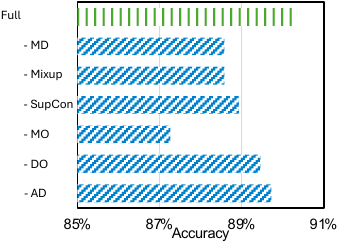}
    \caption{Ablation study, in which MD stands for modality disentanglement, SupCon refers to supervised contrastive loss, MO and DO represents orthogonal loss in modality and domain and AD means adversarial disentanglement}
    \label{fig:ablationStudy}
    \vspace{-20pt}
\end{minipage}
\end{wrapfigure}
To investigate the source of \name's generalization capability further, 
Figure~\ref{fig:tSNE_feature} visualizes the feature representations used by the classifiers for prediction. 
Among all models, \name yields the smallest gap between source and target domains, suggesting that it effectively captures domain-invariant features.

Grad-CAM\cite{CV_selvaraju_ICCV17} highlights the most influential image regions contributing to a model’s decision. 
To provide additional evidence that \name captures domain-invariant features, 
Figure~\ref{fig:cam} compares the Grad-CAM visualizations of \name and the strongest baseline, CLMC. 
As shown, \name focuses on causal, semantically meaningful regions associated with disaster-related concepts, 
while being less influenced by spurious factors such as background textures or irrelevant objects.
\subsection{Ablation Study}
Figure \ref{fig:ablationStudy} presents the average classification accuracy on the target domain for task (1) of CrisisMMD, showing the performance impact of removing each component from \name. A noticeable drop in accuracy occurs whenever a key component is excluded, highlighting the contribution of each module to the overall effectiveness of the framework.

We find that the unified representation across modalities and the mixup operation based on this representation contribute most to generalization performance.
The orthogonal regularization further enhances disentanglement across both modalities and domains, leading to more robust feature separation.
\vspace{-10pt}
\section{Conclusion}
\label{sec:Conclusion}
In this work, we presented \name, a causality-guided multimodal domain generalization framework for disaster classification on social media.
Through adversarial disentanglement and unified representation learning, \name captures domain-invariant causal features, reduces reliance on spurious correlations, and enables the seamless adaptation of single-modality DG techniques to the multimodal setting.
Experiments on CrisisMMD and DMD show that \name achieves an improvement of 4\% to 21\% over strong baselines.
By strengthening the robustness of crisis monitoring models, \name has the potential to support more reliable and timely decision-making in real-world disaster response.
\vspace{-10pt}
\bibliographystyle{splncs04} 
\bibliography{ref/ref} 

@inproceedings{MMDG_dong_NIPS23,
    title={Sim{MMDG}: A Simple and Effective Framework for Multi-modal Domain Generalization},
    author={Dong, Hao and Nejjar, Ismail and Sun, Han and Chatzi, Eleni and Fink, Olga},
    booktitle={Advances in Neural Information Processing Systems (NeurIPS)},
    year={2023}
}

@inproceedings{MMDG_peng_CVPR22,
  title	= {Balanced Multimodal Learning via On-the-fly Gradient Modulation},
  author = {Peng, Xiaokang and Wei, Yake and Deng, Andong and Wang, Dong and Hu, Di},
  booktitle	= {Proceedings of the IEEE/CVF Conference on Computer Vision and Pattern Recognition},
  year	= {2022}
}

@inproceedings{MMDG_fan_NIPS24,
 author = {Fan, Yunfeng and Xu, Wenchao and Wang, Haohao and Guo, Song},
 booktitle = {Advances in Neural Information Processing Systems},
 doi = {10.52202/079017-2133},
 editor = {A. Globerson and L. Mackey and D. Belgrave and A. Fan and U. Paquet and J. Tomczak and C. Zhang},
 pages = {66773--66795},
 publisher = {Curran Associates, Inc.},
 title = {Cross-modal Representation Flattening for Multi-modal Domain Generalization},
 url = {https://proceedings.neurips.cc/paper_files/paper/2024/file/7b5d3047939b63ed97fcbbee23f8eb77-Paper-Conference.pdf},
 volume = {37},
 year = {2024}
}

@inproceedings{MMDG_hai_ICCV25,
  title={Bridging domain generalization to multimodal domain generalization via unified representations},
  author={Huang, Hai and Xia, Yan and Zhou, Sashuai and Wang, Hanting and Wang, Shulei and Zhao, Zhou},
  booktitle={Proceedings of the IEEE/CVF International Conference on Computer Vision},
  pages={22488--22498},
  year={2025}
}

@inproceedings{DG_paras_WSDM24,
author = {Sheth, Paras and Moraffah, Raha and Kumarage, Tharindu S. and Chadha, Aman and Liu, Huan},
title = {Causality Guided Disentanglement for Cross-Platform Hate Speech Detection},
year = {2024},
isbn = {9798400703713},
publisher = {Association for Computing Machinery},
address = {New York, NY, USA},
url = {https://doi.org/10.1145/3616855.3635771},
doi = {10.1145/3616855.3635771},
booktitle = {Proceedings of the 17th ACM International Conference on Web Search and Data Mining},
pages = {626–635},
numpages = {10},
location = {Merida, Mexico},
series = {WSDM '24}
}

@inproceedings{DG_fangrui_CVPR22,
  title={Causality inspired representation learning for domain generalization},
  author={Lv, Fangrui and Liang, Jian and Li, Shuang and Zang, Bin and Liu, Chi Harold and Wang, Ziteng and Liu, Di},
  booktitle={Proceedings of the IEEE/CVF conference on computer vision and pattern recognition},
  pages={8046--8056},
  year={2022}
}

@inproceedings{DG_siyin_LREC_COLING24,
  title={Domain generalization via causal adjustment for cross-domain sentiment analysis},
  author={Wang, Siyin and Zhou, Jie and Chen, Qin and Zhang, Qi and Gui, Tao and Huang, Xuan-Jing},
  booktitle={Proceedings of the 2024 Joint International Conference on Computational Linguistics, Language Resources and Evaluation (LREC-COLING 2024)},
  pages={5286--5298},
  year={2024}
}

@article{Crisis_xie_IPM26,
  title={Advancing cross-domain emergency classification with multi-view adversarial learning},
  author={Xie, Yuhan and Lyu, Chen and Qu, Zheng and Liu, Chunmei},
  journal={Information Processing \& Management},
  volume={63},
  number={2},
  pages={104442},
  year={2026},
  publisher={Elsevier}
}

@inproceedings{Crisis_mouzannar_ISCRAM18,
  title={Damage Identification in Social Media Posts using Multimodal Deep Learning.},
  author={Mouzannar, Hussein and Rizk, Yara and Awad, Mariette},
  booktitle={ISCRAM},
  year={2018},
  organization={Rochester, NY, USA}
}

@inproceedings{Crisis_abavisani_CVPR2020,
  title={Multimodal categorization of crisis events in social media},
  author={Abavisani, Mahdi and Wu, Liwei and Hu, Shengli and Tetreault, Joel and Jaimes, Alejandro},
  booktitle={Proceedings of the IEEE/CVF conference on computer vision and pattern recognition},
  pages={14679--14689},
  year={2020}
}

@inproceedings{Crisis_Mandal_WWW24,
author = {Mandal, Bishwas and Khanal, Sarthak and Caragea, Doina},
title = {Contrastive Learning for Multimodal Classification of Crisis related Tweets},
year = {2024},
isbn = {9798400701719},
publisher = {Association for Computing Machinery},
address = {New York, NY, USA},
url = {https://doi.org/10.1145/3589334.3648143},
doi = {10.1145/3589334.3648143},
booktitle = {Proceedings of the ACM Web Conference 2024},
pages = {4555–4564},
numpages = {10},
location = {Singapore, Singapore},
series = {WWW '24}
}

@inproceedings{Crisis_jain24_ICIC3S24,
  title={Image tweet classification for crisis informative task},
  author={Jain, Tarun and Gopalani, Dinesh and Meena, Yogesh Kumar},
  booktitle={2024 International Conference on Integrated Circuits, Communication, and Computing Systems (ICIC3S)},
  volume={1},
  pages={1--6},
  year={2024},
  organization={IEEE}
}

@inproceedings{Crisis_etal_EMNLP23,
    title = "Natural Disaster Tweets Classification Using Multimodal Data",
    author = "Basit, Mohammad  and
      Alam, Bashir  and
      Fatima, Zubaida  and
      Shaikh, Salman",
    editor = "Bouamor, Houda  and
      Pino, Juan  and
      Bali, Kalika",
    booktitle = "Proceedings of the 2023 Conference on Empirical Methods in Natural Language Processing",
    month = dec,
    year = "2023",
    address = "Singapore",
    publisher = "Association for Computational Linguistics",
    url = "https://aclanthology.org/2023.emnlp-main.471/",
    doi = "10.18653/v1/2023.emnlp-main.471",
    pages = "7584--7594"
}

@ARTICLE{Crisis_Rezk_Access23,
  author={Rezk, Mariham and Elmadany, Noureldin and Hamad, Radwa K. and Badran, Ehab F.},
  journal={IEEE Access}, 
  title={Categorizing Crises From Social Media Feeds via Multimodal Channel Attention}, 
  year={2023},
  volume={11},
  number={},
  pages={72037-72049},
  doi={10.1109/ACCESS.2023.3294474}}

@inproceedings{Crisis_alam_AAAI18,
  title={Crisismmd: Multimodal twitter datasets from natural disasters},
  author={Alam, Firoj and Ofli, Ferda and Imran, Muhammad},
  booktitle={Proceedings of the international AAAI conference on web and social media},
  volume={12},
  number={1},
  year={2018}
}

@article{ML_khosla_NIPS_20,
  title={Supervised contrastive learning},
  author={Khosla, Prannay and Teterwak, Piotr and Wang, Chen and Sarna, Aaron and Tian, Yonglong and Isola, Phillip and Maschinot, Aaron and Liu, Ce and Krishnan, Dilip},
  journal={Advances in neural information processing systems},
  volume={33},
  pages={18661--18673},
  year={2020}
}

@inproceedings{CV_selvaraju_ICCV17,
  title={Grad-cam: Visual explanations from deep networks via gradient-based localization},
  author={Selvaraju, Ramprasaath R and Cogswell, Michael and Das, Abhishek and Vedantam, Ramakrishna and Parikh, Devi and Batra, Dhruv},
  booktitle={Proceedings of the IEEE international conference on computer vision},
  pages={618--626},
  year={2017}
}

@article{CV_simonyan_ICLR15,
  title={Very deep convolutional networks for large-scale image recognition},
  author={Simonyan, Karen and Zisserman, Andrew},
  journal={arXiv preprint arXiv:1409.1556},
  year={2014}
}

@inproceedings{NLP_devlin_NAACL19,
  title={Bert: Pre-training of deep bidirectional transformers for language understanding},
  author={Devlin, Jacob and Chang, Ming-Wei and Lee, Kenton and Toutanova, Kristina},
  booktitle={Proceedings of the 2019 conference of the North American chapter of the association for computational linguistics: human language technologies, volume 1 (long and short papers)},
  pages={4171--4186},
  year={2019}
}

@inproceedings{MM_radford_ICML21,
  title={Learning transferable visual models from natural language supervision},
  author={Radford, Alec and Kim, Jong Wook and Hallacy, Chris and Ramesh, Aditya and Goh, Gabriel and Agarwal, Sandhini and Sastry, Girish and Askell, Amanda and Mishkin, Pamela and Clark, Jack and others},
  booktitle={International conference on machine learning},
  pages={8748--8763},
  year={2021},
  organization={PmLR}
}
\end{document}